\pdfoutput=1

\documentclass[11pt]{article}

\usepackage[]{acl}

\usepackage{times}
\usepackage{latexsym}

\usepackage[T1]{fontenc}

\usepackage[utf8]{inputenc}

\usepackage{microtype}

\usepackage{algorithmicx}  
\usepackage{algpseudocode}  
\usepackage{amsmath}  
\usepackage{color,xcolor}
\usepackage{algorithm} 
\usepackage{pifont}
\usepackage{bm}
\usepackage{amsfonts}
\usepackage{graphicx}
\usepackage[normalem]{ulem}
\usepackage{multirow}
\usepackage{url}
\usepackage{tikz}
\usepackage{xcolor}
\usepackage{tcolorbox}
\usepackage{color,xcolor}
\usepackage{amssymb}
\usepackage{amsopn}
\usepackage{graphicx}
\usepackage{multirow}
\usepackage[english]{babel}
\usepackage{bm}
\usepackage{array}
\usepackage{setspace}
\usepackage{todonotes}
\usepackage{xspace}
\usepackage{amsfonts}
\usepackage{array,multirow}
\usepackage{pgfplots}
\usepackage{tikz}
\usepackage{verbatim}
\usepackage{subfig}
\usepackage{arydshln}
\usepackage{booktabs}

\title{DualNER: A Dual-Teaching framework for Zero-shot Cross-lingual Named Entity Recognition}

\author{Jiali Zeng$^{1}\thanks{\ \ Corresponding author.}$, \ Yufan Jiang$^{1}$, \  Yongjing Yin$^{2}$, \ Xu Wang$^{1}$, \ Binghuai Lin$^{1}$, \ Yunbo Cao$^{1}$ \\
$^{1}$Tencent Cloud Xiaowei, Beijing, China \\
$^{2}$Zhejiang University, Westlake University, Zhejiang, China\\
{\tt \{lemonzeng,garyyfjiang,noorawang,binghuailin,yunbocao\}@tencent.com} \\
{\tt yinyongjing@westlake.edu.cn} \\ 
}

\begin{document}

\maketitle

\begin{abstract}

We present DualNER, a simple and effective framework to make full use of both annotated source language corpus and unlabeled target language text for zero-shot cross-lingual named entity recognition (NER).
In particular, we combine two complementary learning paradigms of NER, i.e., sequence labeling and span prediction, into a unified multi-task framework.
After obtaining a sufficient NER model trained on the source data, we further train it on the target data in a {\it dual-teaching} manner, in which the pseudo-labels for one task are constructed from the prediction of the other task.
Moreover, based on the span prediction, an entity-aware regularization is proposed to enhance the intrinsic cross-lingual alignment between the same entities in different languages.
Experiments and analysis demonstrate the effectiveness of our DualNER.
Code is available at https://github.com/lemon0830/dualNER.

\end{abstract}

\section{Introduction}

Aiming at classifying entities in un-structured text into pre-defined categories, named entity recognition (NER) is an indispensable component for various downstream neural language processing applications such as information retrieval \cite{banerjee-etal-2019-information} and question answering \cite{fabbri-etal-2020-template}.
Current supervised methods have achieved great success
with sufficient manually labeled data,
but 
the fact remains that most of the annotated data are constructed for high-resource languages like English and Chinese, posing a big challenge to low-resource scenarios \cite{mayhew-etal-2017-cheap,bari-etal-2021-uxla}.


To address this issue,
zero-shot cross-lingual NER is proposed to transfer knowledge of NER from high-resource languages to low-resource languages.
The knowledge can be acquired in either of the following two ways: 
1) from aligned cross-lingual word representations or multilingual pre-trained encoder fine-tuned on high-resource languages
\cite{conneau-etal-2020-unsupervised,bari-etal-2021-uxla}.
2) from translated target language data with label projection \cite{mayhew-etal-2017-cheap,jain-etal-2019-entity,liu-etal-2021-mulda}.
These two kinds of methods can be unified into a knowledge distillation (KD) framework, to further improve the cross-lingual NER performance \cite{wu-etal-2020-unitrans,fu-etal-2022-dual}.
Though widely used, the transfer process still suffers from poor translation quality, label projection error and  over-fitting of large-scale multilingual language models. 

In this paper, we present a simple and effective framework, named DualNER, alleviating the above problems from a different angle.
We combine two popular complementary learning paradigms of NER, sequence labeling and span prediction, into a single framework.
Specifically, we first train a teacher NER model by jointly exploiting sequence labeling and span prediction with the annotated source language corpus. 
Unlike the previous KD-based methods
that produce pseudo labels for the corresponding paradigms,
we propose a dual-teaching strategy to make the two paradigms complement each other.
More concretely, the model prediction
for sequence labeling is used to construct the pseudo-labels for span prediction and vice versa.
Furthermore, we propose a multilingual entity-aware regularization forcing same entities in different languages to have similar representations.
By doing this, the trained model is able to leverage the intrinsic cross-lingual alignment across different languages to enhance the cross-lingual transfer ability.

Experiments and analysis conducted on
XTREME for 40 target languages well validate the effectiveness of DualNER.

\begin{figure}[!t]
\centering
\includegraphics[width=1.0\linewidth]{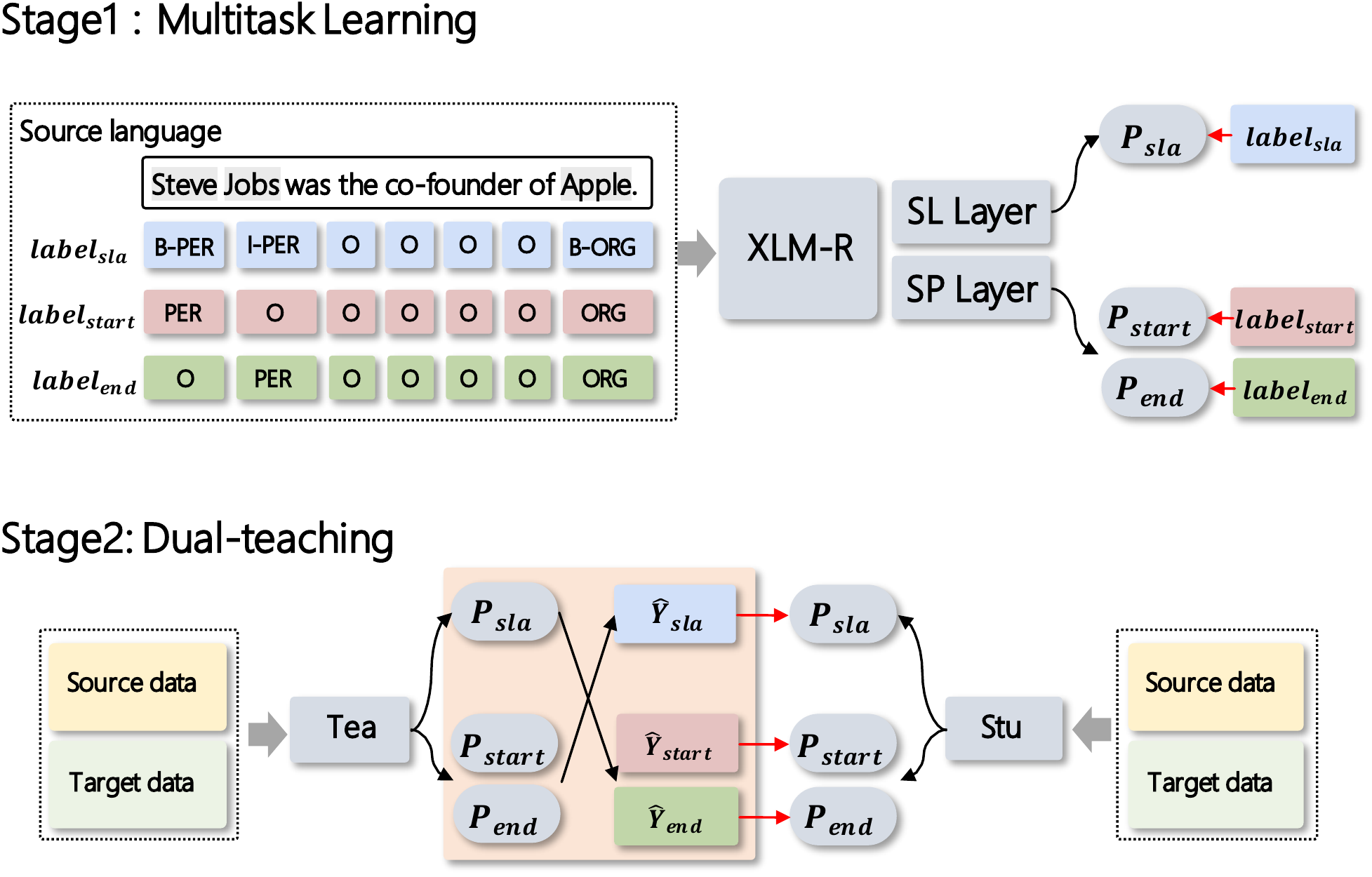}
\caption{
The overall framework of DualNER. SL denotes {\it Sequence Labeling} and SP denotes {\it Span Prediction}.
}
\label{fig_model}
\end{figure}

\section{Framework}

Recently, the dominate paradigm for NER shifts from {\it sequence labeling} \cite{ma-hovy-2016-end-to-end,lample-et-al-2016-neural-architectures,devlin-et-al-2019-bert,xia-et-al-2019-multi,luo-et-al-2020-hierarchical,lin-et-al-2020-triggerNER} to {\it span-level prediction} \cite{jiang-etal-2020-generalizing,ouchi-etal-2020-instance,li-etal-2020-unified,xue-etal-2020-coarse,fu-etal-2021-spanner}.
We combine these two formulations into a unified multitask framework for complementarity.
As shown in Figure \ref{fig_model}, our DualNER consists of three major modules: {\it Token Representation Layer}, {\it Sequence Labeling Layer}, and {\it Span Prediction Layer}.

\subsection{Model}

Given an example of training data $(X, Y_{sla})$, where $X$=$\{x_1, x_i, ..., x_n\}$ is the input sequence and $Y_{sla}$=$\{y_1, y_i, ..., y_n\}$ is the corresponding label (e.g., ``B-ORG'', ``I-PER'', ``O'') sequence,
we can extract the start and end index sequence, $Y_{start}$ and $Y_{end}$, as reference for span prediction, and convert the training instance to a quadruple $(X, Y_{sla}, Y_{start}, Y_{end})$.


\paragraph{Token Representation Layer.}

For the input sequence $X$, we use a multilingual pretrained language models (PLM), e.g., XLM-R, to obtain the contextualized representations $H$=\{$h_1, ..., h_i, ..., h_n$\}.



\paragraph{Sequence Labeling Layer.}

Formally, we stack a softmax classifier layer on $H$, and the objective of sequence labeling is
\begin{equation}
    \mathcal{J}_{sla} = -log(P_{sla}(Y_{sla}|H; \theta, \theta_{sla})),
\end{equation}
where $\theta$ and $\theta_{sla}$ denote the parameters of PLM and the classifier respectively.


\paragraph{Span Prediction Layer.}
For the formulation of span prediction, we adopt two $(C+1)$-class classifiers, where $C$ denotes the number of NER entities (e.g., LOC, PER, ORG, 3 entities in XTREME-40 dataset), 
and one is used to predict whether each token is the start of an entity, and the other is used to predict whether each token is the end.
Formally, given the representations $H$ and two label sequences $Y_{start}$ and $Y_{end}$ of length $n$, 
the losses for start and end index predictions are defined as:
\begin{align}
    \mathcal{J}_{start}&=-log(P_{start}(Y_{start}|H;\theta, \theta_{start})) \\
     \mathcal{J}_{end}&=-log(P_{end}(Y_{end}|H;\theta,\theta_{end})).
\end{align}

\subsection{Training}
To achieve zero-shot cross-language NER, 
we adopt a two-stage training strategy.


\paragraph{Stage 1: Multitask Learning.}
At the first stage, we fine-tune a multilingual pre-trained model on the labeled source language data in a multi-task manner:
\begin{equation}
    \mathcal{J}^{src} = \mathcal{J}^{src}_{sla} + \mathcal{J}^{src}_{start} + \mathcal{J}^{src}_{end}.
\end{equation}



\paragraph{Stage 2: Dual-teaching.}

At the stage two, we focus on generating pseudo labels for both labeled and unlabeled data with the trained NER model $\theta^{tea}$.
In particular, the pseudo labels for the sequence labeling task are converted by the model prediction for the span prediction task, and vice versa.
Specifically,
based on the predictions $P_{sla}$, $P_{start}$ and $P_{end}$ of an input sequence $X^{src}$ (or $X^{trg})$,
we construct the pseudo labels for sequence labeling and span prediction as follows:
\begin{align}
    &\hat{Y}_{sla} = {\rm Sequential} (P_{start}, P_{end}) \\
    &\hat{Y}_{start}, \hat{Y}_{end} = {\rm ExtractSpan} (P_{sla}),
\end{align}
where {\it Sequential} and {\it ExtractSpan} are the corresponding transformation between sequence labels and span labels.


As a result, $X^{src}$ is paired with six label sequences $\{Y^{src}_{sla}, Y^{src}_{start}, Y^{src}_{end}, \hat{Y}^{src}_{sla}, \hat{Y}^{src}_{start}, \hat{Y}^{src}_{end}\}$, and $X^{trg}$ is paired with three pseudo label sequences $\hat{Y}^{trg}_{sla}$, $\hat{Y}^{trg}_{start}$, and $\hat{Y}^{trg}_{end}\}$.
Using the constructed data, we train a student model $\theta^{stu}$ initialized with $\theta^{tea}$ with the following objective:
\begin{align}
    \mathcal{J}^{src} &= 0.5 * \mathcal{J}^{src}(X^{src}, Y^{src}_{sla}, Y^{src}_{start}, Y^{src}_{end}) \\ \notag
    &+ 0.5 * \mathcal{J}^{src}(X^{src}, \hat{Y}^{src}_{sla}, \hat{Y}^{src}_{start}, \hat{Y}^{src}_{end}) \\
    \mathcal{J}^{trg} &= \mathcal{J}^{trg}(X^{trg}, \hat{Y}^{trg}_{sla}, \hat{Y}^{trg}_{start}, \hat{Y}^{trg}_{end}).
\end{align}


\begin{table*}[!ht]
\centering
\small
\begin{spacing}{1.2}
\resizebox{\textwidth}{42mm}{
\begin{tabular}{lcccccccccccccccccccc}
\toprule
\multicolumn{21}{c}{\textit{XLM-R$_{base}$}} \\
\midrule
{\bf Method} & en & af & ar & bg & bn & de & el & es & et & eu & fa & fi & fr & he & hl & hu & id & it & ja & jv \\
CLA & 81.7 & 75.3 & 49.4 & \colorbox{lightgray}{78.2} & 70.2 & \colorbox{lightgray}{74.4} & 74.7 & 69.5 & \colorbox{lightgray}{71.5} & 58.4 & \colorbox{lightgray}{50.6} & \colorbox{lightgray}{75.6} & 75.3 & \colorbox{lightgray}{52.2} & 68.1 & \colorbox{lightgray}{77.0} & \colorbox{lightgray}{47.7} & 77.2 & 21.5 & 54.8 \\
SPAN & \colorbox{lightgray}{83.1} & \colorbox{lightgray}{75.8} & \colorbox{lightgray}{51.0} & 77.8 & \colorbox{lightgray}{70.7} & 74.4 & \colorbox{lightgray}{75.4} & \colorbox{pink}{\bf 80.6} & 67.8 & \colorbox{lightgray}{61.8} & 44.5 & 75.3 & \colorbox{lightgray}{79.7} & 50.3 & \colorbox{lightgray}{68.8} & 76.6 & 47.4 & \colorbox{lightgray}{77.9} & \colorbox{lightgray}{25.5} & \colorbox{lightgray}{57.4} \\
MLT & 82.9 & 75.1 & \colorbox{pink}{\bf 79.3} & 77.6 & 71.3 & 73.6 & 75.2 & 74.3 & 67.8 & 59.8 & 46.6 & 75.3 & 76.5 & 49.7 & 68.3 & 75.1 & 48.8 & 77.0 & 20.1 & 54.2 \\
FILTER & 83.3 & 78.7 & 56.2 & 83.3 & 75.4 & 79.0 & 79.7 & 75.6 & 80.0 & 67.0 & \colorbox{pink}{\bf 70.3} & 80.1 & 79.6 & 55.0 & 72.3 & 80.2 & \colorbox{pink}{\bf 52.7} & \colorbox{pink}{\bf 81.6} & 25.2 & 61.8 \\
DualNER \\
 \ \ \ \ $+$TRG$_{Trans}$ & \colorbox{pink}{\bf 84.1} & 78.0 & 59.0 & 80.0 & \colorbox{pink}{\bf 75.5} & 78.1 & 74.7 & 74.5 & 77.1 & 62.4 & 49.8 & 78.6 & 78.9 & 55.7 &  73.3 & 79.5 & 51.0 & {80.0} & 33.8 & 56.4 \\
 \ \ \ \ $+$TRG$_{Gold}$ & 83.6 & \colorbox{pink}{\bf 80.0} & 62.4 & \colorbox{pink}{\bf 84.6} & 73.9 & \colorbox{pink}{\bf 81.9} & \colorbox{pink}{\bf 80.8} & 79.1 & \colorbox{pink}{\bf 80.4} & \colorbox{pink}{\bf 68.0} & 66.4 & \colorbox{pink}{\bf 82.9} & \colorbox{pink}{\bf 81.8} & \colorbox{pink}{\bf 69.6} & \colorbox{pink}{\bf 74.3} & \colorbox{pink}{\bf 84.5} & {52.5} & {80.3} & \colorbox{pink}{\bf 38.4} & \colorbox{pink}{\bf 63.5} \\

\midrule

{\bf Method} & ka & kk & ko & ml & mr & ms & my & nl & pt & ru & sw & ta & te & th & tl & tr & ur & vi & yo & zh \\
CLA & \colorbox{lightgray}{66.1} & 42.3 & \colorbox{lightgray}{49.3} & \colorbox{lightgray}{59.3} & \colorbox{lightgray}{61.3} & \colorbox{lightgray}{67.2} & \colorbox{lightgray}{55.2} & 79.8 & 77.0 & \colorbox{lightgray}{64.3} & 67.9 & \colorbox{lightgray}{54.1} & 34.0 & \colorbox{lightgray}{0.04} & 69.7 & 76.7 & 56.0 & \colorbox{lightgray}{68.5} & 37.5 & 26.3 \\
SPAN & 65.8 & \colorbox{lightgray}{47.8} & 47.7 & 57.4 & 59.4 & 54.3 & 43.0 & \colorbox{lightgray}{80.9} & \colorbox{lightgray}{80.2} & 61.4 & \colorbox{lightgray}{70.7} & 53.8 & \colorbox{lightgray}{47.1} & 0.03 & \colorbox{lightgray}{72.3} & \colorbox{lightgray}{79.3} & \colorbox{lightgray}{66.3} & 66.8 & \colorbox{pink}{\bf 55.8} & \colorbox{lightgray}{31.9} \\
MLT & 64.5 & 42.8 & 46.7 & 56.3 & 58.8 & 62.3 & 44.2 & 80.5 & 79.3 & 60.8 & 67.6 & 54.6 & 46.2 & 0.01 & 73.0 & 78.1 & 52.9 & 64.9 & 50.2 & 28.4 \\
FILTER & 70.0 & 50.6 & \colorbox{pink}{\bf 63.8} & 67.3 & 66.4 & 68.1 & 60.7 & 83.7 & 81.8 & 71.5 & 68.0 & 62.8 & 56.2 & \colorbox{pink}{\bf 1.5} & 74.5 & 80.9 & 71.2 & 76.2 & 40.4 & 35.9 \\
DualNER \\
 \ \ \ \ $+$TRG$_{Trans}$ & 71.7 & 50.0 & 55.7 & 69.4 & 63.6 & \colorbox{pink}{\bf 70.8} & \colorbox{pink}{\bf 67.7} & 83.1 & 80.3 & 63.3 & 70.5 & 58.8 & 50.7 & 0.08 & 78.0 & 81.6 & 63.7 & 73.4 & {37.9} & 43.6 \\
 \ \ \ \ $+$TRG$_{Gold}$ & \colorbox{pink}{\bf 74.1} & \colorbox{pink}{\bf 52.2} & {62.7} & \colorbox{pink}{\bf 71.1} & \colorbox{pink}{\bf 72.2} & 70.4 & 66.2 & \colorbox{pink}{\bf 84.6} & \colorbox{pink}{\bf 85.2} & \colorbox{pink}{\bf 64.8} & \colorbox{pink}{\bf 70.7} & \colorbox{pink}{\bf 66.5} & \colorbox{pink}{\bf 61.6} & {0.1} & \colorbox{pink}{\bf 80.2} & \colorbox{pink}{\bf 86.0} & \colorbox{pink}{\bf 84.3} & \colorbox{pink}{\bf 77.2} & 35.6 & \colorbox{pink}{\bf 46.1} \\

\bottomrule
\bottomrule

\multicolumn{21}{c}{\textit{InfoXLM$_{large}$}} \\
\midrule
{\bf Method} & en & af & ar & bg & bn & de & el & es & et & eu & fa & fi & fr & he & hl & hu & id & it & ja & jv \\
CLA & 84.5 & 81.4 & \colorbox{lightgray}{64.2} & 83.2 & \colorbox{lightgray}{78.3} & 81.2 & 81.5 & \colorbox{lightgray}{77.6} & \colorbox{lightgray}{79.0} & \colorbox{lightgray}{67.0} & \colorbox{lightgray}{68.9} & 80.9 & 80.1 & \colorbox{lightgray}{58.0} & 74.1 & \colorbox{lightgray}{82.0} & \colorbox{lightgray}{61.0} & 82.4 & 28.0 & \colorbox{lightgray}{61.2} \\
SPAN & \colorbox{lightgray}{85.6} & \colorbox{lightgray}{81.7} & 64.2 & \colorbox{lightgray}{84.6} & 77.3 & \colorbox{lightgray}{81.9} & \colorbox{lightgray}{81.9} & 74.3 & 76.6 & 66.4 & 61.2 & \colorbox{lightgray}{82.2} & \colorbox{lightgray}{80.3} & 54.9 & \colorbox{lightgray}{75.1} & 81.4 & 56.8 & \colorbox{lightgray}{82.4} & \colorbox{lightgray}{31.9} & 59.0 \\
MLT & 85.4 & 83.5 & 62.5 & 84.1 & 78.0 & 80.9 & 82.9 & 80.5 & 78.1 & \colorbox{pink}{\bf 68.3} & 62.6 & 82.4 & 81.6 & 61.2 & 74.8 & 82.5 & 55.8 & 82.4 & 29.9 & 63.9 \\
DualNER \\
 \ \ \ \ $+$TRG$_{Trans}$ & 85.7 & \colorbox{pink}{\bf 83.2} & 65.6 & 84.1 & 80.3 & \colorbox{pink}{\bf 82.8} & 80.9 & 74.5 & \colorbox{pink}{\bf 82.6} & 61.7 & 67.3 & \colorbox{pink}{\bf 82.7} & 82.1 & 66.3 & 77.7 & 82.3 & \colorbox{pink}{\bf 68.1} & \colorbox{pink}{\bf 83.9} & 52.8 & \colorbox{pink}{\bf 68.3} \\
 \ \ \ \ $+$TRG$_{Gold}$ & \colorbox{pink}{\bf 85.9} & 82.5 & \colorbox{pink}{\bf 73.5} & \colorbox{pink}{\bf 86.5} & \colorbox{pink}{\bf 82.9} & 81.8 & \colorbox{pink}{\bf 83.0} & \colorbox{pink}{\bf 83.2} & 79.4 & {67.8} & \colorbox{pink}{\bf 71.9} & 82.3 & \colorbox{pink}{\bf 86.3} & \colorbox{pink}{\bf 74.9} & \colorbox{pink}{\bf 79.5} & \colorbox{pink}{\bf 85.3} & {55.1} & 83.7 & \colorbox{pink}{\bf 55.1} & 68.2 \\

\midrule

{\bf Method} & ka & kk & ko & ml & mr & ms & my & nl & pt & ru & sw & ta & te & th & tl & tr & ur & vi & yo & zh \\
CLA & \colorbox{lightgray}{71.7} & 59.9 & \colorbox{lightgray}{60.6} & \colorbox{lightgray}{65.7} & 68.5 & 71.1 & \colorbox{lightgray}{51.8} & 85.4 & 83.3 & 72.0 & 71.6 & \colorbox{lightgray}{61.5} & \colorbox{lightgray}{57.1} & 0.01 & 75.0 & 83.8 & \colorbox{lightgray}{78.2} & 76.6 & 36.4 & 35.0 \\
SPAN & 71.2 & \colorbox{pink}{60.2} & 57.8 & 64.2 & \colorbox{lightgray}{70.1} & \colorbox{lightgray}{73.3} & 43.9 & \colorbox{lightgray}{86.2} & \colorbox{lightgray}{83.8} & \colorbox{lightgray}{73.6} & \colorbox{lightgray}{72.5} & 60.0 & 51.5 & \colorbox{lightgray}{0.02} & \colorbox{lightgray}{78.0} & \colorbox{lightgray}{85.7} & 75.1 & \colorbox{lightgray}{80.3} & \colorbox{pink}{\bf 56.2} & \colorbox{lightgray}{38.9} \\
MLT & 74.7 & {56.1} & 61.6 & 68.5 & 67.9 & 72.1 & 46.2 & 86.0 & 83.1 & 74.2 & 72.8 & 61.3 & 54.1 & 0.02 & 76.3 & 84.6 & 75.6 & 78.1 & 43.8 & 36.7 \\
DualNER \\
 \ \ \ \ $+$TRG$_{Trans}$ & 78.6 & 56.1 & 69.3 & 72.6 & \colorbox{pink}{\bf 71.5} & 72.1 & \colorbox{pink}{\bf 68.6} & 86.3 & 85.0 & 72.4 & \colorbox{pink}{\bf 73.3} & 66.2 & 56.5 & 0.05 & \colorbox{pink}{\bf 80.6} & 83.2 & 80.2 & \colorbox{pink}{\bf 83.1} & 42.3 & 55.5 \\
 \ \ \ \ $+$TRG$_{Gold}$ & \colorbox{pink}{\bf 79.0} & 51.5 & \colorbox{pink}{\bf 74.2} & \colorbox{pink}{\bf 75.3} & 68.6 & \colorbox{pink}{\bf 74.0} & 66.0 & \colorbox{pink}{\bf 86.7} & \colorbox{pink}{\bf 86.5} & \colorbox{pink}{\bf 80.4} & 71.4 & \colorbox{pink}{\bf 73.6} & \colorbox{pink}{\bf 62.4} & \colorbox{pink}{\bf 0.08} & 76.9 & \colorbox{pink}{\bf 85.9} & \colorbox{pink}{\bf 87.5} & 82.3 & 42.9 & \colorbox{pink}{\bf 57.3} \\
\bottomrule
\end{tabular}
}
\end{spacing}
\caption{
\label{tab_results_main}
{\bf Experimental results on the test sets of XTREME-40 NER}. We highlight better results between {\it CLA} and {\it SPAN} with \colorbox{lightgray}{gray} and highlight best results among all methods with \colorbox{pink}{pink}.
}
\end{table*}

Furthermore, in order to strengthen the correlation of the same entities across languages, we present an entity-aware regularization term. 
We illustrate an example in Appendix \ref{sec:appendix}.
More concretely, for the $j$-th entity, we extract the start token and the end token by applying {\it argmax} to the distributions $P_{start}$ and $P_{end}$, and obtain its representation $r_j$ by concatenating the representations of the two tokens.
We use a mean square error (MSE) loss to pull the representations of the same entities across different languages together:
\begin{equation}
    \mathcal{J}_{mse} = - \frac{1}{C} \sum^{C}_{c=1} \frac{1}{|R_c|} \sum_{(r_m, r_q) \in R_c, m \neq q} (r_m - r_q)^2, \label{equation_regularization}
\end{equation}
where $C$ is the number of NER entities and $R_c$ is the representation set of the $c$-th entity in a mini-batch.

The overall training objective is defined as:
\begin{equation}
    \mathcal{J} = \mathcal{J}^{src} + \mathcal{J}^{trg} + \alpha \cdot  \mathcal{J}_{mse},
\end{equation}
where $\alpha$ is a hyper-parameter to balance the effect of MSE loss.
During training, we update the teacher NER model $\theta^{tea}$ using the better student model $\theta^{stu}$ based on the validation performance.
At inference time, we only use the prediction of {\it Span Prediction Layer}.

\section{Experiments \& Analysis}

\subsection{Setup}

The proposed method is evaluated on the cross-lingual NER dataset from the XTREME-40 benchmark \cite{hu-etal-2020-xtreme}.
Named entities in Wikipedia are annotated with LOC, PER, and ORG tags in BOI-2 format.
We try two types of unlabeled target language data: {\bf Natural Language Text}, 
the target language text in the training set of XTREME-40;
and {\bf Translation Text}
\cite{fang-etal-2021-filter}.
We take XLM-R-base \cite{conneau-etal-2020-unsupervised} and InfoXLM-large \cite{chi-etal-2021-infoxlm} as our backbones, and set $\alpha$ as 0.5.
Detailed experimental setups are shown in Appendix \ref{sec:setting}.
We use entity-level F1-score of all language development sets to choose the best checkpoint, and report the F1-score on each test set of each language.

\subsection{Main Result}
We compare DualNER to the following baselines: 1) {\it FILTER} \cite{fang-etal-2021-filter}, which feeds paired language input into PLM and is trained with self-teaching;
2) {\it CLA}, which formulates NER as a sequence labeling problem;
3) {\it SPAN}, which formulates NER as a span prediction problem;
and 4) {\it MLT}, the model trained after our Stage 1.
Besides, we name DualNER trained on unlabeled target natural language text as {\it DualNER+TRG$_{Gold}$}, while denote DualNER trained on target language translation text as {\it DualNER+TRG$_{Trans}$}.

Table \ref{tab_results_main} reports the zero-shot cross-lingual NER results.
The conclusions are as follows: 1) 
{\it CLA} and {\it SPAN}
have no obvious advantages over each other.
2) {\it DualNER} significantly outperforms the baselines on almost all of the languages, demonstrating the effectiveness of our proposed method.
3) Directly combining {\it CLA} and {\it SPAN} into a multitask learning framework (i.e., {\it MLT}) fails to achieve consistent improvement.
This observation shows that the gain of DualNER entirely comes from the proposed {dual-teaching} training strategy, rather than the usage of multitask learning.
4) As expected, using natural language text (i.e., {\it DualNER+TRG$_{Gold}$}) achieves better performance compared to translation text (i.e., {\it DualNER+TRG$_{Trans}$}), since translations possibly lose the idiomatic expressions of some entities.

\subsection{Ablation Study}

To analyze the impact of different components of DualNER, we investigate the following three variants: 1) {\it DualNER w/o $\mathcal{J}_{mse}$}, removing the entity-aware regularization; 2) {\it DualNER w/ selfKL}, where {\it Dual-teaching} is replaced by {\it Self-teaching} with KL loss at the Stage 2. 
3) {\it DualNER w/o TRG}, where we only use the source language data in the Stage 2. 
We take XLM-R$_{base}$ as the backbone.
The results are listed in Table \ref{tab_results_ablation_study}.
Compared with {\it DualNER w/ selfKL}, {\it DualNER} obtains a significant improvement of 3.46 points, validating our motivation in making use of complementarity of different task paradigms of NER.
The degradation of {\it DualNER w/o $\mathcal{J}_{mse}$} and {\it DualNER w/o TRG} confirm the intrinsic cross-lingual alignment and the importance of task-related target language information.

\begin{table}[!t]
\centering
\small
\begin{tabular}{lc}
\toprule
{\bf Model} & {\bf F1} \\
\midrule
MLT & 61.53$\pm$0.53 \\
DualNER+TRG$_{Gold}$ & {\bf 68.64}$\pm$0.06 \\
\ \ \ \ w/o $\mathcal{J}_{mse}$ & 68.05$\pm$0.49 \\
\ \ \ \ w/ selfKL & 65.18$\pm$0.65 \\
\ \ \ \ w/o TRG & 62.17$\pm$0.52 \\
\bottomrule
\end{tabular}
\caption{
\label{tab_results_ablation_study}
{\bf Ablation Study}.
We run 3 times with different random seeds and report mean and standard deviation on all the validation sets.
}
\end{table}

\begin{figure}[!th]
\centering
\includegraphics[width=0.8\linewidth]{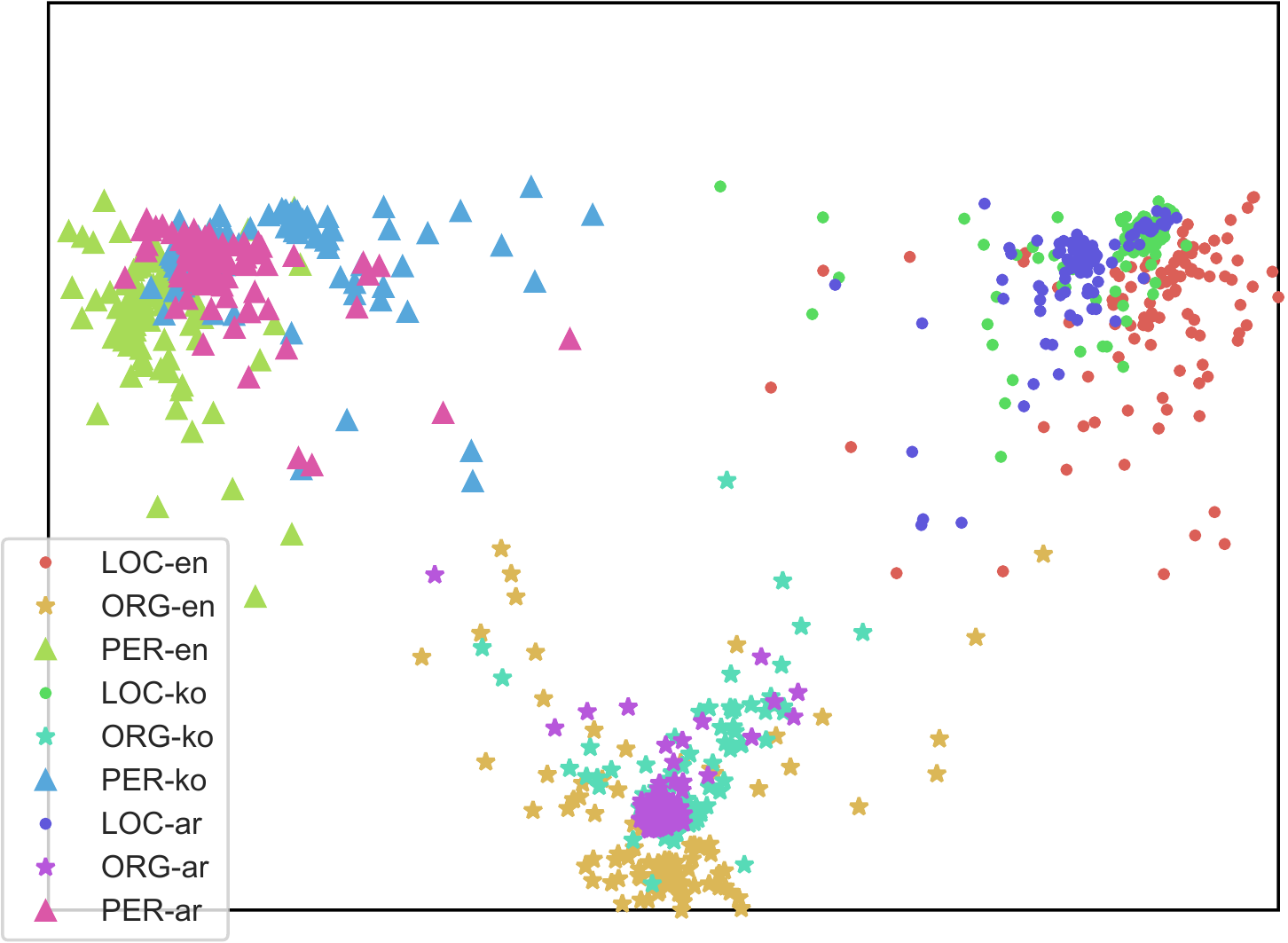}
\caption{
The visualization of the entity
representations in different languages, where the triangle-shaped,
circle-shaped, and
pentagonal-shaped(blue) points denote location,
organization, and person entities, respectively.
}
\label{fig_entity}
\end{figure}

\subsection{Visualization}

We choose English, Korean, and Arabic, which comes from different language families, and visualize the entity representations $r$ in Eq. \ref{equation_regularization} with hypertools \cite{heusser-etal-2017-hypertools}.
As shown in Figure \ref{fig_entity}, the representations of different entities in the same language are clearly distributed in different regions, while the representations of the same entity across different languages are concentrated.



\begin{figure}[!t]
  \centering
  \begin{tikzpicture}[scale = 0.85]
    \footnotesize{
      \begin{axis}[
      ymajorgrids,
  xmajorgrids,
  grid style=dashed,
      width=.50\textwidth,
      height=.30\textwidth,
      legend style={at={(0.20,0.12)}, anchor=south west},
      xlabel={\scriptsize{Corpus size}},
      ylabel={\scriptsize{F1}},
      ylabel style={yshift=-1em},
      xlabel style={yshift=0.0em},
      ymin=45,ymax=70, ytick={45, 50, 55, 60, 65, 70},
      xmin=18,xmax=100,xtick={20, 40, 60, 80, 100},
      legend style={yshift=2pt, legend plot pos=right, legend columns=3 ,font=\scriptsize,cells={anchor=west}}
      ]

      \addplot[red!70,mark=diamond*,line width=1pt] coordinates {(20,63.52) (40,64.51) (60,64.92) (80,68.44) (100,68.65)};
      \addlegendentry{\scriptsize SampleSrc}
      \addplot[blue!70,mark=diamond*,line width=1pt] coordinates {(20,49.49) (40,57.19) (60,59.27) (80,60.46) (100,60.88)};
      \addlegendentry{\scriptsize MLT}
      \end{axis}
     }
  \end{tikzpicture}
  \caption{{\bf Effect of Source Language Corpus Size.} We report F1-score on all the validation sets.}\label{fig_corpus_size}
\end{figure}
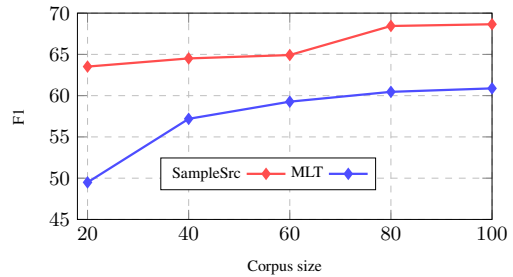

\subsection{Effect of Source Language Corpus Size}
In this experiment, we study the impact of annotated source language corpus size on DualNER by sampling different percentages of annotated source language corpus for the Stage 1.
Meanwhile, we remove the labels of the remaining source data, and mix it with the unlabeled target language text for the Stage 2.
Figure \ref{fig_corpus_size} shows the comparison between {\it DualNER} and {\it MLT}.
Surprisingly, DualNER trained with only 20\% of annotated source data achieves better performance than MLT trained using complete data, demonstrating the data-efficiency of our proposed method.

\section{Conclusion}
In this paper, we propose a simple and effective dual-teaching framework, coined DualNER, for zero-shot cross-lingual named entity recognition.
In particular, DualNER makes full use of the exchangeability of the labels in span prediction and sequence labeling, and generates abundant pseudo data for available labeled and unlabeled data.
Experiments and analysis validate the effectiveness of our DualNER.

\section{Limitations}
The performance of DualNER relies on the capability of cross-lingual transfer of multilingual pretrained models.
In practice, for an adequate quality of the pseudo-labels generated in the stage 2, it is necessary to ensure that the NER model has acquired certain ability to conduct cross-lingual transfer in the stage 1.


\bibliography{anthology,custom}

\begin{thebibliography}{24}
\expandafter\ifx\csname natexlab\endcsname\relax\def\natexlab#1{#1}\fi

\bibitem[{Banerjee et~al.(2019)Banerjee, Chakraborty, Tripathi, Gupta, and
  Kumar}]{banerjee-etal-2019-information}
Partha~Sarathy Banerjee, Baisakhi Chakraborty, Deepak Tripathi, Hardik Gupta,
  and Sourabh~S. Kumar. 2019.
\newblock \href {https://doi.org/10.1007/s11277-019-06501-z} {A information
  retrieval based on question and answering and {NER} for unstructured
  information without using {SQL}}.
\newblock \emph{Wirel. Pers. Commun.}, 108(3):1909--1931.

\bibitem[{Bari et~al.(2021)Bari, Mohiuddin, and Joty}]{bari-etal-2021-uxla}
M~Saiful Bari, Tasnim Mohiuddin, and Shafiq Joty. 2021.
\newblock \href {https://doi.org/10.18653/v1/2021.acl-long.154} {{UXLA}: A
  robust unsupervised data augmentation framework for zero-resource
  cross-lingual {NLP}}.
\newblock In \emph{Proceedings of the 59th Annual Meeting of the Association
  for Computational Linguistics and the 11th International Joint Conference on
  Natural Language Processing (Volume 1: Long Papers)}, pages 1978--1992,
  Online. Association for Computational Linguistics.

\bibitem[{Chi et~al.(2021)Chi, Dong, Wei, Yang, Singhal, Wang, Song, Mao,
  Huang, and Zhou}]{chi-etal-2021-infoxlm}
Zewen Chi, Li~Dong, Furu Wei, Nan Yang, Saksham Singhal, Wenhui Wang, Xia Song,
  Xian-Ling Mao, Heyan Huang, and Ming Zhou. 2021.
\newblock \href {https://doi.org/10.18653/v1/2021.naacl-main.280} {{I}nfo{XLM}:
  An information-theoretic framework for cross-lingual language model
  pre-training}.
\newblock In \emph{Proceedings of the 2021 Conference of the North American
  Chapter of the Association for Computational Linguistics: Human Language
  Technologies}, pages 3576--3588, Online. Association for Computational
  Linguistics.

\bibitem[{Conneau et~al.(2020)Conneau, Khandelwal, Goyal, Chaudhary, Wenzek,
  Guzm{\'a}n, Grave, Ott, Zettlemoyer, and
  Stoyanov}]{conneau-etal-2020-unsupervised}
Alexis Conneau, Kartikay Khandelwal, Naman Goyal, Vishrav Chaudhary, Guillaume
  Wenzek, Francisco Guzm{\'a}n, Edouard Grave, Myle Ott, Luke Zettlemoyer, and
  Veselin Stoyanov. 2020.
\newblock \href {https://doi.org/10.18653/v1/2020.acl-main.747} {Unsupervised
  cross-lingual representation learning at scale}.
\newblock In \emph{Proceedings of the 58th Annual Meeting of the Association
  for Computational Linguistics}, pages 8440--8451, Online. Association for
  Computational Linguistics.

\bibitem[{Devlin et~al.(2019)Devlin, Chang, Lee, and
  Toutanova}]{devlin-et-al-2019-bert}
Jacob Devlin, Ming{-}Wei Chang, Kenton Lee, and Kristina Toutanova. 2019.
\newblock \href {https://doi.org/10.18653/v1/n19-1423} {{BERT:} pre-training of
  deep bidirectional transformers for language understanding}.
\newblock In \emph{Proceedings of the 2019 Conference of the North American
  Chapter of the Association for Computational Linguistics: Human Language
  Technologies, {NAACL-HLT} 2019, Minneapolis, MN, USA, June 2-7, 2019, Volume
  1 (Long and Short Papers)}, pages 4171--4186. Association for Computational
  Linguistics.

\bibitem[{Fabbri et~al.(2020)Fabbri, Ng, Wang, Nallapati, and
  Xiang}]{fabbri-etal-2020-template}
Alexander Fabbri, Patrick Ng, Zhiguo Wang, Ramesh Nallapati, and Bing Xiang.
  2020.
\newblock \href {https://doi.org/10.18653/v1/2020.acl-main.413} {Template-based
  question generation from retrieved sentences for improved unsupervised
  question answering}.
\newblock In \emph{Proceedings of the 58th Annual Meeting of the Association
  for Computational Linguistics}, pages 4508--4513, Online. Association for
  Computational Linguistics.

\bibitem[{Fang et~al.(2021)Fang, Wang, Gan, Sun, and
  Liu}]{fang-etal-2021-filter}
Yuwei Fang, Shuohang Wang, Zhe Gan, Siqi Sun, and Jingjing Liu. 2021.
\newblock \href {https://ojs.aaai.org/index.php/AAAI/article/view/17512}
  {{FILTER:} an enhanced fusion method for cross-lingual language
  understanding}.
\newblock In \emph{Thirty-Fifth {AAAI} Conference on Artificial Intelligence,
  {AAAI} 2021, Thirty-Third Conference on Innovative Applications of Artificial
  Intelligence, {IAAI} 2021, The Eleventh Symposium on Educational Advances in
  Artificial Intelligence, {EAAI} 2021, Virtual Event, February 2-9, 2021},
  pages 12776--12784. {AAAI} Press.

\bibitem[{Fu et~al.(2021)Fu, Huang, and Liu}]{fu-etal-2021-spanner}
Jinlan Fu, Xuanjing Huang, and Pengfei Liu. 2021.
\newblock \href {https://doi.org/10.18653/v1/2021.acl-long.558} {{S}pan{NER}:
  Named entity re-/recognition as span prediction}.
\newblock In \emph{Proceedings of the 59th Annual Meeting of the Association
  for Computational Linguistics and the 11th International Joint Conference on
  Natural Language Processing (Volume 1: Long Papers)}, pages 7183--7195,
  Online. Association for Computational Linguistics.

\bibitem[{Fu et~al.(2022)Fu, Lin, Yang, and Jiang}]{fu-etal-2022-dual}
Yingwen Fu, Nankai Lin, Ziyu Yang, and Shengyi Jiang. 2022.
\newblock \href {https://doi.org/10.48550/arXiv.2204.00796} {A dual-contrastive
  framework for low-resource cross-lingual named entity recognition}.
\newblock \emph{CoRR}, abs/2204.00796.

\bibitem[{Heusser et~al.(2017)Heusser, Ziman, Owen, and
  Manning}]{heusser-etal-2017-hypertools}
Andrew~C. Heusser, Kirsten Ziman, Lucy L.~W. Owen, and Jeremy~R. Manning. 2017.
\newblock \href {http://jmlr.org/papers/v18/17-434.html} {Hypertools: a python
  toolbox for gaining geometric insights into high-dimensional data}.
\newblock \emph{J. Mach. Learn. Res.}, 18:152:1--152:6.

\bibitem[{Hu et~al.(2020)Hu, Ruder, Siddhant, Neubig, Firat, and
  Johnson}]{hu-etal-2020-xtreme}
Junjie Hu, Sebastian Ruder, Aditya Siddhant, Graham Neubig, Orhan Firat, and
  Melvin Johnson. 2020.
\newblock \href {http://proceedings.mlr.press/v119/hu20b.html} {{XTREME:} {A}
  massively multilingual multi-task benchmark for evaluating cross-lingual
  generalisation}.
\newblock In \emph{Proceedings of the 37th International Conference on Machine
  Learning, {ICML} 2020, 13-18 July 2020, Virtual Event}, volume 119 of
  \emph{Proceedings of Machine Learning Research}, pages 4411--4421. {PMLR}.

\bibitem[{Jain et~al.(2019)Jain, Paranjape, and Lipton}]{jain-etal-2019-entity}
Alankar Jain, Bhargavi Paranjape, and Zachary~C. Lipton. 2019.
\newblock \href {https://doi.org/10.18653/v1/D19-1100} {Entity projection via
  machine translation for cross-lingual {NER}}.
\newblock In \emph{Proceedings of the 2019 Conference on Empirical Methods in
  Natural Language Processing and the 9th International Joint Conference on
  Natural Language Processing (EMNLP-IJCNLP)}, pages 1083--1092, Hong Kong,
  China. Association for Computational Linguistics.

\bibitem[{Jiang et~al.(2020)Jiang, Xu, Araki, and
  Neubig}]{jiang-etal-2020-generalizing}
Zhengbao Jiang, Wei Xu, Jun Araki, and Graham Neubig. 2020.
\newblock \href {https://doi.org/10.18653/v1/2020.acl-main.192} {Generalizing
  natural language analysis through span-relation representations}.
\newblock In \emph{Proceedings of the 58th Annual Meeting of the Association
  for Computational Linguistics}, pages 2120--2133, Online. Association for
  Computational Linguistics.

\bibitem[{Lample et~al.(2016)Lample, Ballesteros, Subramanian, Kawakami, and
  Dyer}]{lample-et-al-2016-neural-architectures}
Guillaume Lample, Miguel Ballesteros, Sandeep Subramanian, Kazuya Kawakami, and
  Chris Dyer. 2016.
\newblock \href {https://doi.org/10.18653/v1/n16-1030} {Neural architectures
  for named entity recognition}.
\newblock In \emph{{NAACL} {HLT} 2016, The 2016 Conference of the North
  American Chapter of the Association for Computational Linguistics: Human
  Language Technologies, San Diego California, USA, June 12-17, 2016}, pages
  260--270. The Association for Computational Linguistics.

\bibitem[{Li et~al.(2020)Li, Feng, Meng, Han, Wu, and
  Li}]{li-etal-2020-unified}
Xiaoya Li, Jingrong Feng, Yuxian Meng, Qinghong Han, Fei Wu, and Jiwei Li.
  2020.
\newblock \href {https://doi.org/10.18653/v1/2020.acl-main.519} {A unified
  {MRC} framework for named entity recognition}.
\newblock In \emph{Proceedings of the 58th Annual Meeting of the Association
  for Computational Linguistics}, pages 5849--5859, Online. Association for
  Computational Linguistics.

\bibitem[{Lin et~al.(2020)Lin, Lee, Shen, Moreno, Huang, Shiralkar, and
  Ren}]{lin-et-al-2020-triggerNER}
Bill~Yuchen Lin, Dong{-}Ho Lee, Ming Shen, Ryan Moreno, Xiao Huang, Prashant
  Shiralkar, and Xiang Ren. 2020.
\newblock \href {https://doi.org/10.18653/v1/2020.acl-main.752} {Triggerner:
  Learning with entity triggers as explanations for named entity recognition}.
\newblock In \emph{Proceedings of the 58th Annual Meeting of the Association
  for Computational Linguistics, {ACL} 2020, Online, July 5-10, 2020}, pages
  8503--8511. Association for Computational Linguistics.

\bibitem[{Liu et~al.(2021)Liu, Ding, Bing, Joty, Si, and
  Miao}]{liu-etal-2021-mulda}
Linlin Liu, Bosheng Ding, Lidong Bing, Shafiq Joty, Luo Si, and Chunyan Miao.
  2021.
\newblock \href {https://doi.org/10.18653/v1/2021.acl-long.453} {{M}ul{DA}: A
  multilingual data augmentation framework for low-resource cross-lingual
  {NER}}.
\newblock In \emph{Proceedings of the 59th Annual Meeting of the Association
  for Computational Linguistics and the 11th International Joint Conference on
  Natural Language Processing (Volume 1: Long Papers)}, pages 5834--5846,
  Online. Association for Computational Linguistics.

\bibitem[{Luo et~al.(2020)Luo, Xiao, and Zhao}]{luo-et-al-2020-hierarchical}
Ying Luo, Fengshun Xiao, and Hai Zhao. 2020.
\newblock \href {https://ojs.aaai.org/index.php/AAAI/article/view/6363}
  {Hierarchical contextualized representation for named entity recognition}.
\newblock In \emph{The Thirty-Fourth {AAAI} Conference on Artificial
  Intelligence, {AAAI} 2020, The Thirty-Second Innovative Applications of
  Artificial Intelligence Conference, {IAAI} 2020, The Tenth {AAAI} Symposium
  on Educational Advances in Artificial Intelligence, {EAAI} 2020, New York,
  NY, USA, February 7-12, 2020}, pages 8441--8448. {AAAI} Press.

\bibitem[{Ma and Hovy(2016)}]{ma-hovy-2016-end-to-end}
Xuezhe Ma and Eduard~H. Hovy. 2016.
\newblock \href {https://doi.org/10.18653/v1/p16-1101} {End-to-end sequence
  labeling via bi-directional lstm-cnns-crf}.
\newblock In \emph{Proceedings of the 54th Annual Meeting of the Association
  for Computational Linguistics, {ACL} 2016, August 7-12, 2016, Berlin,
  Germany, Volume 1: Long Papers}. The Association for Computer Linguistics.

\bibitem[{Mayhew et~al.(2017)Mayhew, Tsai, and Roth}]{mayhew-etal-2017-cheap}
Stephen Mayhew, Chen-Tse Tsai, and Dan Roth. 2017.
\newblock \href {https://doi.org/10.18653/v1/D17-1269} {Cheap translation for
  cross-lingual named entity recognition}.
\newblock In \emph{Proceedings of the 2017 Conference on Empirical Methods in
  Natural Language Processing}, pages 2536--2545, Copenhagen, Denmark.
  Association for Computational Linguistics.

\bibitem[{Ouchi et~al.(2020)Ouchi, Suzuki, Kobayashi, Yokoi, Kuribayashi,
  Konno, and Inui}]{ouchi-etal-2020-instance}
Hiroki Ouchi, Jun Suzuki, Sosuke Kobayashi, Sho Yokoi, Tatsuki Kuribayashi,
  Ryuto Konno, and Kentaro Inui. 2020.
\newblock \href {https://doi.org/10.18653/v1/2020.acl-main.575} {Instance-based
  learning of span representations: A case study through named entity
  recognition}.
\newblock In \emph{Proceedings of the 58th Annual Meeting of the Association
  for Computational Linguistics}, pages 6452--6459, Online. Association for
  Computational Linguistics.

\bibitem[{Wu et~al.(2020)Wu, Lin, Karlsson, Huang, and
  Lou}]{wu-etal-2020-unitrans}
Qianhui Wu, Zijia Lin, B{\"{o}}rje~F. Karlsson, Biqing Huang, and Jianguang
  Lou. 2020.
\newblock \href {https://doi.org/10.24963/ijcai.2020/543} {Unitrans : Unifying
  model transfer and data transfer for cross-lingual named entity recognition
  with unlabeled data}.
\newblock In \emph{Proceedings of the Twenty-Ninth International Joint
  Conference on Artificial Intelligence, {IJCAI} 2020}, pages 3926--3932.
  ijcai.org.

\bibitem[{Xia et~al.(2019)Xia, Zhang, Yang, Li, Du, Wu, Fan, Ma, and
  Yu}]{xia-et-al-2019-multi}
Congying Xia, Chenwei Zhang, Tao Yang, Yaliang Li, Nan Du, Xian Wu, Wei Fan,
  Fenglong Ma, and Philip~S. Yu. 2019.
\newblock \href {https://doi.org/10.18653/v1/p19-1138} {Multi-grained named
  entity recognition}.
\newblock In \emph{Proceedings of the 57th Conference of the Association for
  Computational Linguistics, {ACL} 2019, Florence, Italy, July 28- August 2,
  2019, Volume 1: Long Papers}, pages 1430--1440. Association for Computational
  Linguistics.

\bibitem[{Xue et~al.(2020)Xue, Yu, Zhang, Liu, Zhang, and
  Wang}]{xue-etal-2020-coarse}
Mengge Xue, Bowen Yu, Zhenyu Zhang, Tingwen Liu, Yue Zhang, and Bin Wang. 2020.
\newblock \href {https://doi.org/10.18653/v1/2020.emnlp-main.514}
  {Coarse-to-fine pre-training for named entity recognition}.
\newblock In \emph{Proceedings of the 2020 Conference on Empirical Methods in
  Natural Language Processing, {EMNLP} 2020, Online, November 16-20, 2020},
  pages 6345--6354. Association for Computational Linguistics.

\end{thebibliography}

\appendix

\section{Example of Entity-aware Regularization}
\label{sec:appendix}

\begin{figure}[!h]
\centering
\includegraphics[width=1.0\linewidth]{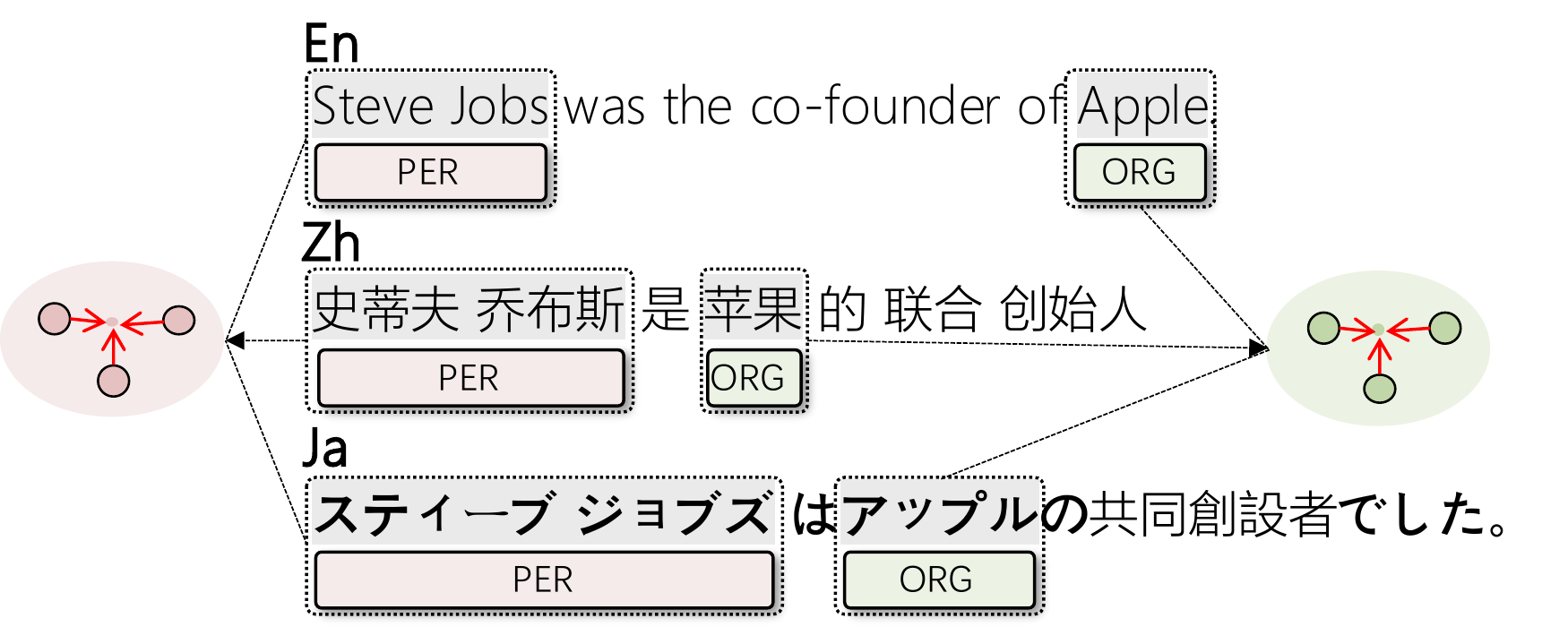}
\caption{
Entity-aware Regularization.
}
\label{fig_regularization}
\end{figure}

Figure \ref{fig_regularization} illustrates an example of entity-aware regularization.

\section{Settings for Different Pretrained
Models} \label{sec:setting}

\begin{table}[!h]
\centering
\small
\begin{spacing}{1.0}
\begin{tabular}{lcccc}
\toprule
{\bf Model} & {\bf Batch Size} & {\bf Epoch} & {\bf Warmup} & {\bf lr} \\
\midrule
\multicolumn{5}{c}{\bf Stage 1} \\
XLM-R$_{base}$ & 128 & 8 & 300 & 2e-5 \\
InfoXLM$_{large}$ & 128 & 8 & 300 & 2e-5 \\
\midrule
\multicolumn{5}{c}{\bf Stage 2} \\
XLM-R$_{base}$ & 500 & 8 & 300 & 2e-5 \\
InfoXLM$_{large}$ & 128 & 8 & 300 & 2e-5 \\
\bottomrule
\end{tabular}
\end{spacing}
\caption{
\label{tab_settings}
Hyper-parameters settings for different pretrained models.
}
\end{table}

In this paper, we fine-tuned different pretrained models including XLM-R-base and InfoXLM-large.
We evaluate the model each 250 steps.
The batch size, training epoch, warmup steps and learning rate in two-stage training are list in Table \ref{tab_settings}.

\end{document}